\documentclass{article}
\usepackage{spconf,amsmath,graphicx}
\usepackage{multicol}
\usepackage{multirow}
\usepackage{arydshln}
\usepackage{xcolor}
\usepackage{lipsum}
\usepackage{microtype}
\usepackage{graphicx}
\usepackage{subfigure}
\usepackage[hidelinks]{hyperref}       
\usepackage{booktabs} 
\usepackage{amsfonts}       
\usepackage{nicefrac}       
\usepackage{arydshln}
\usepackage{algpseudocode}
\usepackage{algorithm}

\usepackage{tikz}
\usepackage{pgfplots}
\title{Evaluating Self-Supervised Speech Representations for Indigenous American Languages}
\name{Chih-Chen Chen$^1$, William Chen$^2$, Rodolfo Zevallos$^3$, John E. Ortega$^4$}
\address{Taipei Medical University$^1$ \\Carnegie Mellon University$^2$ \\  Universitat Pompeu Fabra$^3$ \\ Northeastern University$^4$}
\begin{document}
%
\maketitle
\begin{abstract}
The application of self-supervision to speech representation learning has garnered significant interest in recent years, due to its scalability to large amounts of unlabeled data. However, much progress, both in terms of pre-training and downstream evaluation, has remained concentrated in monolingual models that only consider English. Few models consider other languages, and even fewer consider indigenous ones. In our submission to the New Language Track of the ASRU 2023 ML-SUPERB Challenge, we present an ASR corpus for Quechua, an indigenous South American Language. We benchmark the efficacy of large SSL models on Quechua, along with 6 other indigenous languages such as Guarani and Bribri, on low-resource ASR. Our results show surprisingly strong performance by state-of-the-art SSL models, showing the potential generalizability of large-scale models to real-world data.
\end{abstract}
\section{Introduction}
\label{sec:intro}
In recent years, the fields of Natural Language Processing (NLP) and speech processing has witnessed remarkable advancements, with applications ranging from machine translation and sentiment analysis to voice assistants and chatbots. These developments have predominantly focused on widely spoken languages such as English and Mandarin. However, the vast linguistic diversity represented by indigenous languages across the globe remains largely unexplored in the context of language processing.

Indigenous languages are the ancestral tongues of diverse communities with rich cultural heritage and profound connections to the environment in which they are spoken. These languages often exhibit distinct linguistic characteristics, deviating from the structures and conventions of widely studied languages. By expanding the scope of language processing to include indigenous languages, we can foster linguistic inclusivity and empower indigenous communities to participate in the digital era while preserving their linguistic and cultural identities.

One compelling reason to focus on indigenous languages in language processing is the potential for social impact. Many Indigenous communities face challenges related to limited access to information and technology, which further exacerbate social and economic disparities. By developing NLP models and applications tailored to indigenous languages, we can bridge the digital divide and enable these communities to leverage technology for communication, education, and cultural preservation. Such efforts have the potential to enhance language revitalization efforts, foster inter-generational transmission of knowledge, and promote cultural preservation within indigenous communities.

Our core contribution is a Quechua ASR corpus, which serves as our submission to the New Language Track of the ML-SUPERB Challenge. We evaluate the effectiveness of different self-supervised learning (SSL) models for speech on this corpus, along with other indigenous American languages: Bribri, Guarani, Kotiria, Wa'ikhana, and Totonac.

We first introduce general characteristics of American indigenous languages and discuss the challenges in modelling them due to their unique linguistic natures. We then provide a brief overview of each language to this study, highlighting some key linguistic properties. As our other core contribution, we discuss research in American indigenous languages in both the fields of NLP and speech processing, hoping to create a bridge in the literature for communities. 

\section{American Indigenous Languages}

American Indigenous Languages encompass a diverse range of language families and isolates, each with its own linguistic features. The languages of this region exhibit a remarkable variety of phonological, morphological, syntactic, and semantic structures, reflecting the rich linguistic diversity of the continent. A persistent challenge in modelling these languages, similar to many indigenous languages, is the frequency of code-switching. Coupled with the lack of both linguistic and electronic resources for these languages, creating language technologies for indigenous languages remains a significant challenge despite the exponential progress in NLP and speech processing.

Broadly speaking, the indigenous languages of the Americas are morphologically-rich, often exhibiting agglutinative or polysynthetic structures. These languages tend to have extensive systems of affixation, where morphemes are added to roots to convey meaning and grammatical information. For example, in Quechua or Guarani, complex words can be formed through the addition of numerous affixes to a single root. This makes them particularly challenging for NLP and language modelling tasks, due to the higher frequency of rare words.

\subsection{Quechua}
Quechua is a family of closely related languages spoken by around 10 million people across South America. While primarily spoken in the Andean regions, communities can also be found along the plains and valleys connecting the Amazon to the Pacific. Quechua is considered one of the most widely spoken indigenous language families in the Americas. While there are regional variations, Quechua languages share many common linguistic characteristics. These variations are broadly separated into two distinct categories: Quechua I and Quechua II. The former refers to the varieties of Quechua spoken in the central parts of Peru, while the latter is spoken in Southern Peru, Bolivia, and Colombia. Dialects between the two categories may not necessarily be mutually intelligible, which is what distinguishes Quechua as a language family, and makes it particularly challenging for language processing.

\subsection{Bribri}
Bribri, also known as the Bribri-Poró language, is spoken by the Bribri people of Costa Rica. It belongs to the Chibchan language family, which is primarily found in Central America. The Bribri language specifically falls under the Guaymí subgroup of the Chibchan family. Geographically, the Bribri language is primarily spoken in the Talamanca region of Costa Rica, specifically in the southern parts of Limón and northern parts of Puntarenas provinces. It is a tonal language, meaning that pitch variations can distinguish between different words or meanings. The language also exhibits agglutinative tendencies, where words are formed by adding affixes to a base or root.

\subsection{Guarani}
The Guarani language is an indigenous language spoken by the Guarani people in South America. It is a member of the Tupi-Guarani language family, which encompasses several languages across Brazil, Paraguay, Argentina, and Bolivia. Guarani is one of the most widely spoken indigenous languages in the Americas, with 4-6 million speakers. Guarani is mainly distributed in Paraguay, where it has official status alongside Spanish. It is also spoken in parts of northeastern Argentina, southeastern Bolivia, and southern Brazil. Like many other American languages, Guarani is agglutinative. Guarani is a highly regular language with consistent pronunciation rules, which may be notable for speech processing.

\subsection{Kotiria}
Kotiria, also known as Wanano, is an indigenous language spoken by the Kotiria people who reside in the Vaupés region of Colombia and Brazil. Kotiria is part of the larger Eastern Tukanoan language family, which includes several other indigenous languages spoken in the northwest Amazon region. Kotiria language is primarily spoken in the upper and middle basins of the Vaupés River, which runs through the Amazon rainforest. The language is concentrated in remote areas of the Colombian Vaupés Department and the Brazilian state of Amazonas. Like Bribri, it is both agglutinative and tonal. 

\subsection{Wa'ikhana}
Also known as Cubeo, Wa'ikhana is spoken by the Cubeo people in the northwest Amazon region, primarily in Colombia and Brazil. Wa'ikhana belongs to the Tucanoan language family, which encompasses several indigenous languages spoken in the northwest Amazon. Wa'ikhana is also both agglutinative and tonal.

\subsection{Totonac}
The Totonac language is an indigenous language spoken by the Totonac people in Mexico. It belongs to the Totonacan language family, which is primarily found in the states of Veracruz, Puebla, and parts of Hidalgo in eastern Mexico. Totonac is both agglutinative and tonal.

\section{Indigenous Languages in Language Processing}

\subsection{Community Efforts}

In the field of NLP, several initiatives have been started to encourage further research in indigenous languages. While the majority are workshops for general low-resource NLP \cite{mtsummit-2021-technologies, amta-2022-biennial-association}, newer efforts have also targeted indigenous languages \cite{americasnlp-2021-natural, americasnlp-2023-natural, orife2020masakhane, nekoto-etal-2020-participatory}. For American indigenous languages specifically, the AmericasNLP \cite{americasnlp-2021-natural, americasnlp-2023-natural} community has helped driven research by improving the visibility of authors from indigenous communities. AmericasNLP also hosts an annual shared task, similar to those found in machine and speech translation workshops \cite{ebrahimi-etal-2023-findings, mager-etal-2021-findings}, to further integrate state-of-the-art methods with indigenous languages.

In speech processing, research for indigenous languages is more \textit{ad hoc}, with numerous decentralized efforts from a variety of research groups. Contrary to NLP, indigenous languages play a more common role in SOTA models \cite{chen2023reducing, babu2021xls, radford2022robust, pratap20_interspeech_mls, zhang2023google} and benchmarks \cite{conneau2022fleurs, shi2023exploration, gales2014speech}. Annual challenges, primarily for speech translation, also help bring SOTA methods to these languages \cite{agrawal-etal-2023-findings}.

\subsection{Research for Quechua}
Quechua has enjoyed high amounts of attention in the field of NLP relative to other American languages, likely due to its higher population and speakers and thus larger amount of resources. Much of the early work involved analyzing the morphological properties of Quechua, such as through finite state transducers, \cite{rios-gonzales-castro-mamani-2014-morphological, rios2010applying, rios-gonzales-gohring-2013-machine} and developing toolkits to process the language \cite{rios2015basic, rios2008quechua, rios2011spell}. Neural approaches were first adopted in the context of machine translation \cite{ortega-pillaipakkamnatt-2018-using, ortega2020neural, chen-fazio-2021-morphologically} before being adopted to masked language models \cite{zevallos-etal-2022-introducing}.

Fewer studies have been done on Quechua on the speech processing side, likely due to the lack of available data. To our knowledge, Siminchik \cite{cardenas2018siminchik} was the first speech corpus for Quechua. While the authors established baseline results with HMM-based systems, a full release of the data was never realized. Similarly, Huqariq \cite{zevallos-etal-2022-huqariq} is a multilingual collection of four native Peruvian languages, including Quechua, that has yet to be publicly released. Quechua was featured in two speech processing challenges, the AmericasNLP 2022 Competition and IWSLT 2023 \cite{agrawal-etal-2023-findings}. The latter case saw the first evaluation of Quechua with Transformer-based \cite{vaswani2017attention}, with participants greatly leveraging pre-trained SSL models \cite{e-ortega-etal-2023-quespa}. However, the performance of other SSL methods on Quechua remain an open question, as all participants utilized either XLSR 53 \cite{conneau2020unsupervised} or XLS-R 128 \cite{babu2021xls}.

\section{Corpus} \label{sec:data}

Our submission to the ASRU 2023 ML-SUPERB challenge is derived from the Siminchik corpus \cite{cardenas2018siminchik}, which contains recordings of two different Quechua II dialects. The first is Chanca Quechua, which is spoken primarily in Ayacucho and its surrounding areas in Peru. The other dialect is Collao Quechua, which is spoken in Cusco and Puno.  

Siminchik \cite{cardenas2018siminchik} consists of crowd-resourced transcriptions of radio recordings. The initial recordings were collected from radio channels in Ayachuco and Apurimac for Chanca Quechua, while for Collao Quechua they were collected from Puno and Cusco. Advertisements, music, and segments of Spanish speech were filtered out, yielding 97 hours of audio. Audio clips were then segmented to a maximum of 30 seconds, although beginning and end words were likely trimmed due to the imperfect truncation method \cite{cardenas2018siminchik, giannakopoulos2015pyaudioanalysis}. 

The authors of Siminchik conducted several post-processing steps on the annotated transcripts. First, punctuation was removed and casing was normalized. Due to differences in dialects, interjections were also normalized. This was accomplished by making a dictionary that mapped expressions to a specific word form. Furthermore, dialectal differences also require the spelling of the ASR transcripts to be normalized to a common form. 
This was accomplished using a finite state transducer-based normalization toolkit for Quechua \cite{rios-gonzales-castro-mamani-2014-morphological}, which adheres to the spelling of the Chanca dialect. 

For our submission to ML-SUPERB, we sample 90 minutes of speech from the corpus, to create the 10-minute and 1-hour training set, the 10-minute development set, and the 10-minute test. 

\section{Experimental Setup}

\subsection{Data}

We conduct our experiments on 6 indigenous American languages: Quechua, Bribri, Guarani, Kotiria, Wa'ikhana, and Totonac. For Quechua, we use the data we submitted to the ML-SUPERB Challenge described in Section \ref{sec:data}. For Bribri, Guarani, Kotiria, and Wa'ikhana, we use data from the 2022 AmericasNLP competition. The Totonac data is derived from a study on using speech technologies for endangered languages documentation. Due to the lack of a public testing split for AmericasNLP, we split the data ourselves from the provided validation sets. All data is formatted in the style of ML-SUPERB \cite{shi2023ml}, which consists of a 1-hour training set, a 10-minute training set, a 10-minute validation set, and a 10-minute testing set. This low-resource data setting most accurately benchmarks the capabilities of SSL models for indigenous languages in real-world settings, due to the lack of both available labeled and unlabeled data.
\begin{table*}[htb]
    \centering
    \caption{Evaluation of SSL models on each indigenous language on the 10-minute set, measured in character error rate (CER $\downarrow$).}
    \resizebox {\linewidth} {!} {
\begin{tabular}{lc|cccccc|c}
\toprule
Model & Hours & Quechua & Bribri & Guarani & Kotiria & Wa'ikhana & Totonac & Average \\
\midrule
XLSR 53 & 56k  & 47.8 & 54.6 & 37.6 & 64.2 & 83.3 & 29.6 & 52.9\\
XLS-R 128  & 436k & \textbf{42.5} & \textbf{49.5} & \textbf{27.5} & \textbf{51.2} & \textbf{62.2} & \textbf{27.7} & 43.4\\
mHuBERT & 13.5k & 47.7& 54.3 & 35.2 & 64.8 & 84.8 & 30.1 & 52.8 \\ 
\bottomrule
\end{tabular}
}
    \label{tab:results_10m}
    \vspace{-0.5cm}
\end{table*}

\begin{table*}[htb]
    \centering
    \caption{Evaluation of SSL models on each indigenous language on the 1 hour set, measured in character error rate (CER $\downarrow$).}
    \resizebox {\linewidth} {!} {
\begin{tabular}{lc|cccccc|c}
\toprule
Model & Hours & Quechua & Bribri & Guarani & Kotiria & Wa'ikhana & Totonac & Average \\
\midrule
XLSR 53 & 56k  & 37.5 & 49.5 & 31.5 & 49.9 & 62.4 & 26.0 & 42.8\\
XLS-R 128 & 436k & \textbf{34.0} & \textbf{44.1} & \textbf{24.0} &\textbf{ 43.4 }& \textbf{55.1 }& \textbf{20.6 }& \textbf{36.8} \\
mHuBERT& 13.5k & 37.1 & 49.2 & 32.0 & 50.6 & 62.3 & 26.1 & 42.8\\ 
\bottomrule
\end{tabular}
}
    \label{tab:results_1h}
    \vspace{-0.5cm}
\end{table*}
\subsection{Self-Supervised Models}
We evaluate three SSL models on each language, along with log-Mel filterbank features (FBANK). The models are described as follows:

\subsubsection{XLSR 53}
XLSR 53 \cite{conneau2020unsupervised} is trained on 56k hours of multilingual data for 53 languages, which are pre-dominantly European. It uses the 317M parameter wav2vec architecture \cite{schneider19_wav2vec}, which consists of a convolutional feature extractor and Transformer encoder \cite{vaswani2017attention} trained with contrastive loss.

\subsubsection{XLS-R 128}
XLS-R 128 \cite{babu2021xls} is the large-scale extension of XLSR 53, trained on 436k hours of multilingual data across 128 languages. It instead uses the wav2vec 2.0 \cite{baevskiw2v} architecture, which also includes a convolutional feature extractor and Transformer encoder, but is trained with both contrastive and codebook prediction losss.

\subsubsection{mHuBERT}
mHuBERT \cite{lee-etal-2022-textless} builds off of the HuBERT \cite{hsuHubert} architecture, which uses an iterative approach to SSL. HuBERT models are trained to predict discrete representations of masked speech. After each iteration of pre-training, hidden representations are extracted from the model and clustered using k-means, creating the discrete targets for the next round of pre-training. mHuBERT was trained multilingually on 3 languages: Spanish, French, and Italian, each 4.5k hours of data. It uses the 95M parameter HuBERT Base architecture, which modifies the wav2vec 2.0 design for pure codebook prediction.

\subsection{Training Settings}
We conduct all experiments using the ESPnet \cite{watanabe2018espnet} toolkit with the official settings of the ML-SUPERB competition. The SSL model is used as a frozen feature extractor, such that the hidden representation of each layer is obtained. The layer-wise outputs of combined via a weighted sum, where the weight is learned during training. These outputs are then down projected to a hidden size of 80 and then augmented with SpecAug \cite{park2019specaugment}, before being used as the model inputs of a Transformer encoder \cite{vaswani2017attention}. The Transformer consists of 2 layers, each with a hidden size of 256, 8 attention heads, and a feed-forward size of 1024. Models are trained with CTC loss \cite{graves2006connectionist} and the Adam optimizer \cite{kingma2014adam}, with a constant learning rate of 0.0001. Models are trained for a maximum of 15,000 steps and the 5 best checkpoints are averaged for inference, which is performed with CTC greedy decoding.

\section{Results}

Our experimental results are presented in Tables \ref{tab:results_10m} and \ref{tab:results_1h} for the 10-minute and 1-hour settings respectively. Models are evaluated in character error rate (CER).

Similar to the complete ML-SUPERB benchmark, XLS-R 128 \cite{babu2021xls} obtains the highest overall scores in both data settings. The results presented here are even more distinct: XLS-R 128 outperforms all other models every single task. This suggests the powerful generalizability of large-scale multilingual SSL: all evaluated languages (aside from Guarani) were unseen during pre-training. 

The distance between the other two models, XLSR 53 and mHuBERT, is much smaller, with only a difference of 0.1 average CER on the 10-minute track and no significant difference on the 1-hour track. A strong future research question would be to isolate the cause for the lack of difference, as one would expect the model trained on more languages to generalize better.

Overall, we find the results of our evaluation surprisingly strong. The average CER of XLS-R 128 on the 1 hour set is 36.8, only 6.2 CER higher than its average monolingual score on ML-SUPERB. While further improvements are necessary, this shows that it is possible to shrink the gap between high and low-resource languages with powerful cross-lingual transfer learning, even in cases without large-scale unlabeled data.

\section{Conclusion}
While the recent progress of deep learning in NLP and speech processing has significantly accelerated the development of language technologies, the progress has been unequally distributed. In our submission to the ASRU 2023 ML-SUPERB Challenge, we present an ASR corpus for Quechua, an indigenous South American Language. We are the first benchmark the effectiveness of large-scale speech SSL models on indigenous American languages such as Quechua, which are known to be among the most difficult for NLP. We find surprisingly positive results, showing the impressive generalization ability of large-scale multilingual SSL models on new languages.

\clearpage
\bibliographystyle{IEEEbib}
\bibliography{strings,refs}

\end{document}